# Machine Learning for Visual Navigation of Unmanned Ground Vehicles

Artem A. Lenskiy, Jong-Soo Lee,

University of Ulsan, South Korea

**ABSTRACT**

The use of visual information for the navigation of unmanned ground vehicles in a cross-country environment recently received great attention. However, until now, the use of textural information has been somewhat less effective than color or laser range information. This chapter reviews the recent achievements in cross-country scene segmentation and addresses their shortcomings. It then describes a problem related to classification of high dimensional texture features. Finally, it compares three machine learning algorithms aimed at resolving this problem. The experimental results for each machine learning algorithm with the discussion of comparisons are given at the end of the chapter.

## INTRODUCTION

### Literature overview

The area of autonomous driving on- and off-road vehicles is expanding very rapidly. A great deal of work has been done developing autonomous navigational systems for driving along highways and roads in an urban environment. The autonomous navigation in determined and rigid urban environment with lanes, road markers and boards is relatively easier than the off-road autonomous navigation. In off-road navigation the significantly changing environment with fuzzy or no roads creates a new complexity for navigational issues. Only recently has cross-country navigation received appropriate attention. A good example is The Grand Challenge which was launched by the Defense Advanced Research Projects Agency (DARPA) in 2003. The original goal of the project was to stimulate innovation in unmanned ground vehicle navigation. Two years later an unmanned ground vehicle (UGV) named Stanley was able to navigate a 132-mile long off-road course and complete it in 6 hours 53 minutes (Thrun, et al., 2006).

UGVs are usually equipped with multiple sensors to operate in a variety of cross-country environments (Fig.1). This equipment along with sophisticated algorithms serves to solve navigational problems such as map building, path planning, land mark detection, position estimation and obstacle avoidance. In this chapter we focus on the visual terrain segmentation task. The terrain segmentation allows the robot to detect obstacles and select the optimal path. Based on the information obtained by means of terrain segmentation, the robot is able to avoid unnecessary stops caused by traversable tall patches of grass. The segmentation information also allows adjusting traversal velocity depending on the terrain slippery factors.

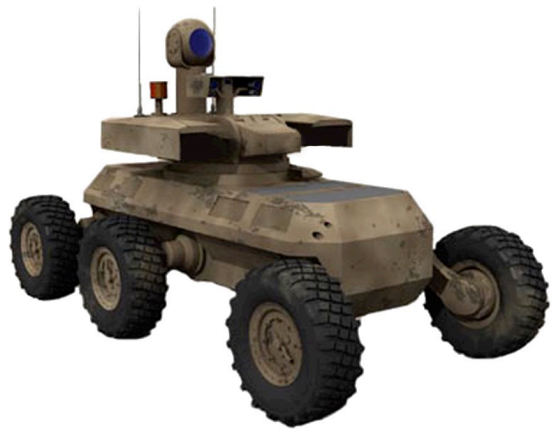

*Figure 1. An unmanned ground vehicle.*

There are multiple ways to segment a cross-country scene image, depending on what image characteristics are taken into account. Regardless of what characteristics are used, the final goal is to separate spatial image regions on the basis of their similarity. In the terrain segmentation task, image characteristics as color(Manduchi, 2006; Rasmussen, 2002), texture(Castano, Manduchi, & Fox, 2001; Sung, Kwak, & Lyou, 2010) and range data(Dahlkamp, Kaehler, Stavens, Thrun, & Bradski, 2006; Lalonde, Vandapel, Huber, & Hebert, 2006) are commonly utilized. The best terrain segmentation results are obtained when all characteristics are incorporated in the segmentation process. Nevertheless, in this chapter texture information is applied for cross-country scene segmentation. Depending on the terrain type, some image characteristics are more distinctive than others. Particularly, color information is useful in distinguishing classes such as sky, dry or green vegetation. However, there are a number of shortcomings associated with color segmentation algorithms. Compared to texture, color based segmentation algorithms are less robust to brightness changes caused fluctuations in natural illumination or shadows. Another demerit is that red, green, and blue color components that constitute color space are less discriminative than multidimensional texture features. Finally, color segmentation does not work at night, while texture segmentation can be applied to IR images captured at night. Nevertheless, adaptive color segmentation algorithms are useful especially in combination with other types of features. The off-road scene segmentation algorithm implemented in Stanley (Dahlkamp, et al., 2006; Thrun, et al., 2006) (the DARPA Grand Challenge winner) did not take into account texture information. There are likely two reasons for this. Texture features are usually computationally expensive to extract, and until now the performance of texture features was quite unsatisfactory compared to other scene characteristics.

Rasmussen (Rasmussen, 2002), provided a comparison of color, texture, distance features measured by the laser range scanner, and their combination for the purpose of cross-country scene segmentation. The segmentation was the worst when texture features were used alone. In the case when 25% of all features were used for training, only 52.3% of the whole feature set was correctly classified. There are two probable explanations of this poor result. One is related to the feature extraction approach. The feature vector consisted of 48 values representing responses of the Gabor filter bank.  Specifically, it consists of 2 phases with 3 wavelengths and 8 equally-spaced orientations. The 48-dimensional vector appears to have enough dimensions to accommodate a wide variety of textures. However, besides the feature dimensionality, the size of texture patches also influence the segmentation quality.  The size of the patch was set to a relatively small constant value equal to 15x15, which led to poor scale invariance. Furthermore, features' locations were calculated on the grid without considering an image content. Another reason of the problematic segmentation results is in the low classifier's capacity. As a classifier, the author used a neural network with only one hidden layer with 20 neurons. A one layer feed-forward neural network is not capable of partitioning concave clusters, while terrain texture features are very irregular.

Sung et al. (Sung, et al., 2010) instead, used a two-layer percpetron with 18 and 12 neurons in the first and second hidden layers correspondingly. The feature vector was composed of the mean and energy values computed for selected sub-bands of two-level Daubechies wavelet transform, resulting in 8 values. These values were calculated for each of three color channels resulting into 24-dimensional feature vector. The experiments were conducted in the following fashion. First, 100 random images from the stored video frames were selected and used to extract training patches. Then among them ten were chosen for testing purposes. The average segmentation rate was 75.1% when two-layer perceptron and 24-dimensional feature vectors were applied. Considering that color information was not explicitly used and only texture features were taken into account, the segmentation rate is promising, although there is still room for improvement. Similarly to Rasmussen (Rasmussen, 2002), the wavelet mean and energy were calculated for fixed 16x16 pixel sub-blocks. Consider a resolution of input images of 720x480 pixel, the sub-block of 16x16 pixels is too small to capture texture characteristics, especially at higher scales, which leads to poor texture scale invariance.

Castano et al. (Castano, et al., 2001) applied Gabor features extracted as described in (Manjunath & Ma, 1996) with 3 scales and 4 orientations. Two statistical texture models were analyzed. The first classifier modeled the probability distribution function of texture features using mixtures of Gaussian and performed a Maximum Likelihood classification. The second classifier represents local statistics by marginal histograms over small image squares. Comparable performances were reached with both models. Particularly, in the case when half of the hand segmented images were used for training and the other half for testing, the classification performance on the cross-country scene images was 70% for mixtures of Gaussian and 66% for histogram based classifiers. Visual analysis of presented segmentation results suggests that the wrong classification happens due to the short range of scale independence of Gabor features.

There are two major directions for algorithmic improvement: features extraction and machine learning. In this chapter we focus on comparison of the following machine learning algorithms: nearest-neighbor algorithm (NNA), multi-layer perceptron(MLP) and support vector machine (SVM). We also analyze the influence of changing the dimension of feature vectors on the quality of feature classification.

**Problem statement**

The majority of papers related to texture segmentation consider homogeneous textures which usually lack real-world problems, when appearance of the same texture greatly changes. Cross-country segmentation brings immense complexity into the texture segmentation task due to high inter- and intraclass variation. For instance, in the example of tree textures, there is a broad variety of trees to consider. Secondly, even for the same type of trees the appearance of their texture patches changes drastically with the changing distance to the camera. When a camera is close enough it is possible to distinguish branches and single leaves so the texture patch has one set of properties; when the camera is further away the tree looks like a green spot, resulting in completely changed properties. The intraclass variation comes from the similar appearance of different texture classes. Textures from different classes may look similar depending on factors such as their distance from the camera, weather conditions and the time of day (fig. 2).

To be able to account for all these possibilities and correctly segment input images, a high-dimensional feature space with a great number of features is needed. That is where data analysis plays a great role. We consider two mutually related machine learning problems. The first one is the generalization problem accounting for transforming the training set in to a more compact and generalized form. The second one is a classification problem; an algorithm learns to predict positions of each class vectors.

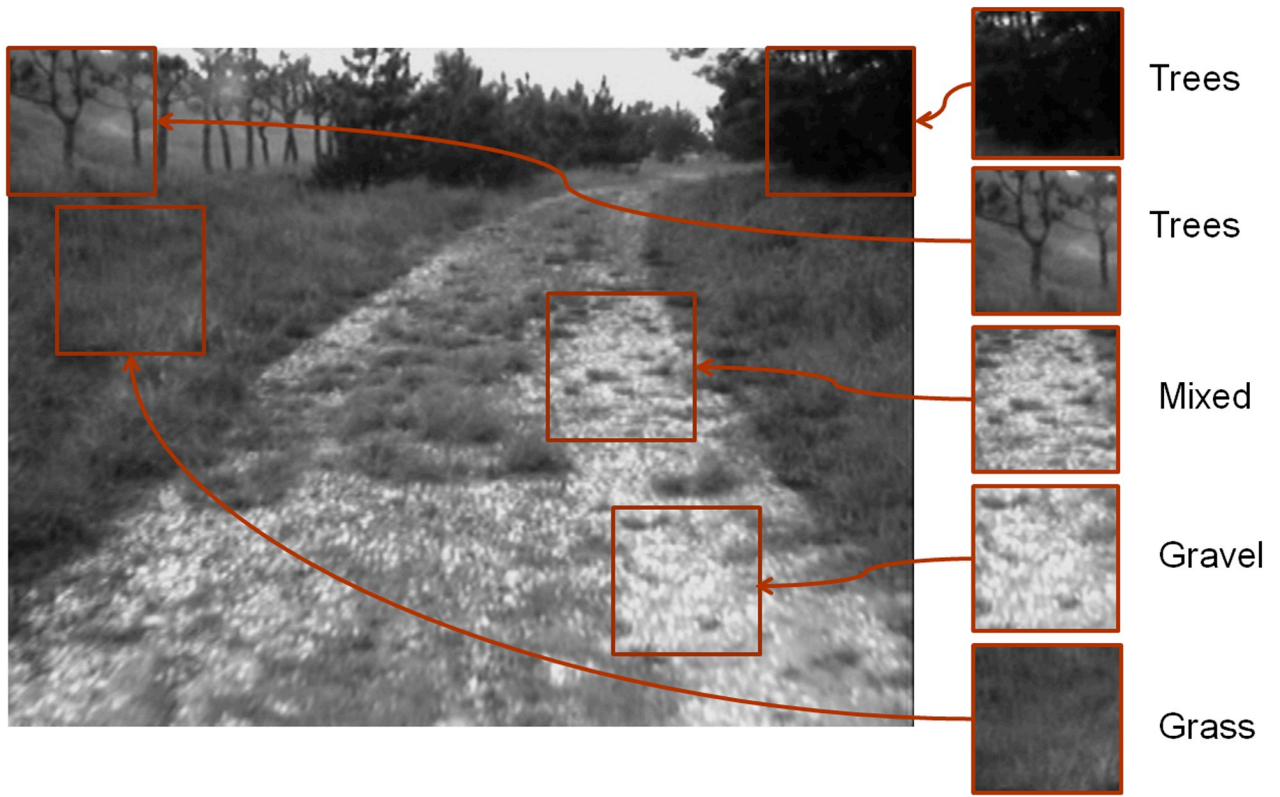

*Figure 2. Texture similarity in between classes and dissimilarity within a class.*

Let $S = \{(x_i, y_i) | i = 1..n\}$ is a training set, where $x \subset \mathbf{R}^d$ is a feature vector and $y = \{1, \ldots, m\}$ is a feature's label, $m$ corresponds to the number of classes. It is often useful to take into account information on how likely a feature was classified to a particular class. Therefore, to allow us to represent non-mutually exclusive classes we code output vectors as $m$ dimensional vector, with $y \subset \mathbf{R}^m$. The goal is to find a transform $f$ such that $f(x) \sim y$ and, in the least mean squares sense we look for a transform minimizing the following loss function:

$$\varepsilon(f) = \sum_{i=1}^{n} (y_i - f(x_i))^2 \qquad (1)$$

It is easy to see that the simplest approach to make the criterion (1) equal to zero is by forcing $f$ to pass through all training samples. Lazy-learning algorithms are good examples of this principle. Although this approach is easy to implement, it has two disadvantages. When a training set contains a great number of samples it becomes time and memory consuming to simply store all samples in the knowledge base. The second disadvantage is related to overfitting of noisy data contained in the training set. On the other hand, inappropriate reduction of the number of training samples would lead to poor generalization and classification. Our goal is to compare and find an appropriate machine learning algorithm which suits a vast amount of high-dimensional vectors by not only minimizing the loss function (1) but also minimizing the computational time and memory demands for classification.

**Cross-country scene segmentation system overview**

Depending on the view point the scene segmentation system can be divided into sub-systems or functioning stages. From the machine learning perspective, the system consists of two stages: the learning stage and the recognition stage. In the learning stage, the training set is transformed into a compact and generalized form suitable for classification. The result of the learning stage is some form of knowledge base which depends on a machine learning algorithm. From the other point of view, the system consists of three subsystems. The first subsystem deals with image preprocessing and texture features extraction. The second subsystem depending on the learning or recognition stage is responsible for supervised learning or features classification. The last subsystem segments the input image using the classification results.

The main focus of this chapter is the learning as well as recognition stages.

**THE TRAINING DATA PREPROCESSING**

Our terrain segmentation system is designed to recognize five different terrain types. The list of terrains includes grass, gravel, trees, dirt/mud and sky. The training data is selected from prerecorded video sequences. The total number of images is 2973, with every 100$^{th}$ being hand segmented. The hand segmentation process itself is a challenge. It is often the case when terrains of different types are mixing up so, it makes difficult to distinguish a region containing only one type of terrain (fig. 2). In this case the region is segmented to the class that pixels are the most represented in the region. Another problem we face during hand segmentation is that terrains residing far away from the camera lack strong textures and are usually blurred; they therefore look similar to one another. In this case very blurred regions are avoided due to their insignificancy for the training set and also due to the fact that the priority of the UGV is to recognize nearest environment rather than distant.

We overall selected and segmented 29 images. Ten of them were selected for the testing purpose and 19 for training. Each training image pair was processed with the subject to extract salient features. Salient features are sorted up into five matrices according to their labels (fig. 3). As for salient features, we chose speeded-up robust features (SURF) (Bay, Ess, Tuytelaars, & Gool,

2008). Each feature consists of two vectors. The first vector contains information of the feature's location, scale and strength, and the second vector is a descriptor characterizing the region around its location.

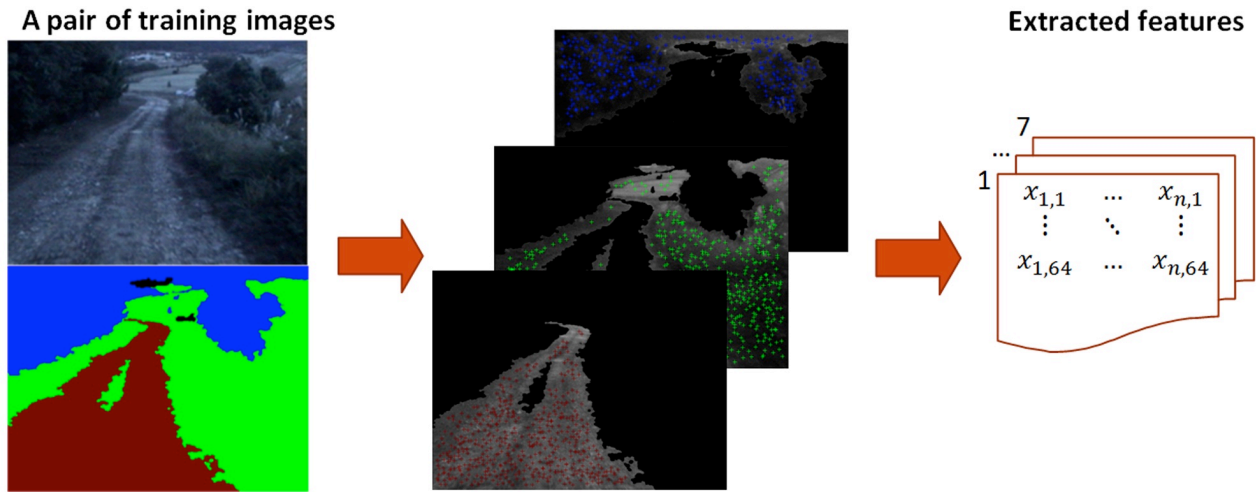

*Figure 3. An schematic representation of features extraction from a training pair.*

The SURF algorithm consists of three stages. In the first stage, interest points and their scales are selected. The features' locations and scales are selected by finding the maxima of the determinant of the Fast-Hessian matrix calculated in scale space. In the second stage, features' orientations are estimated. At this stage Haar wavelet responses are calculated for both *x* and *y* directions surrounding the interest point and the dominant orientation is estimated by calculating the sum of all responses within a sliding orientation window. This direction is then used to create a rotated square around the interest point. Finally, to build the descriptor, an oriented quadratic grid with *n x n* square sub-regions is laid over the interest point. For each square, the vertical $d_y$ and horizontal $d_x$ Haar wavelet responses are computed from 5 x 5 samples. Then, the wavelet responses $d_y$ and $d_x$ and their absolute values $|d_x|$ and $|d_y|$ are summed up over each sub-region forming the description vector.

In our system we experimented with the SURF algorithm as well as with an upright version of the SURF (U-SURF). The latter is a rotation dependent version of the SURF. It skips the second stage and as a result it is faster to compute. We also experimented with two different numbers of sub-regions *n = 4* and *n = 3*. When *n = 4*, the total number of the feature's dimensions is 64, and in the case of *n = 3*, the number of dimensions is 36.

The number of features detected by the SURF algorithm greatly depends on the predefined blob response threshold and the image content. If the threshold is too high than just a few features are detected, if the image is monotonic then the number of detected features is small too. On the other hand if the threshold is low and the image consists of not monotonic regions, then the number of detected features is high. To limit the number of detected features from the top and at the same assure that the number is not too small; we set the blob response threshold to low and then reduce the number of detected features as follows. The first, image is partitioned into boxes with the size of 20 x 20 pixels, then the feature with the highest strength is selected among all features fallen into each box (Figure 4). Therefore, if the image resolution is 640 x 480 pixels and the box size is 20 x 20 pixels, then the maximum number of features equals 768. The advantage of this approach versus those mentioned in the introduction is that the number of features and as well as features' locations are automatically adjusted depending on the image content. Moreover, instead of a fixed window size used in previous approaches, it is automatically adjusted by the SURF algorithm. Extracting features from all of 19 training images results in 11394 labeled features.

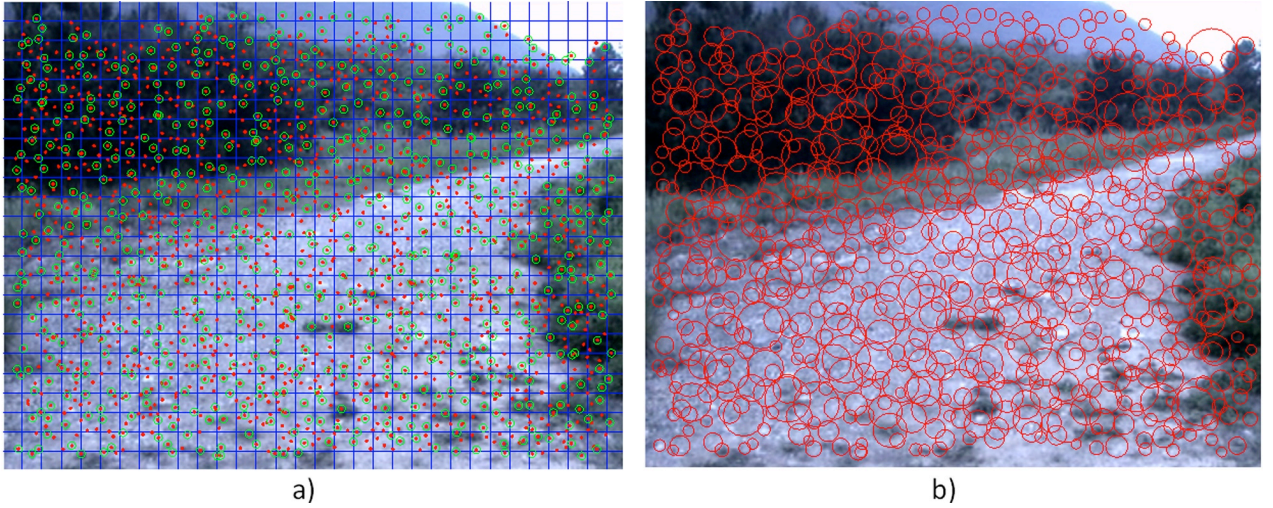

*Figure 4. a) Red points represent centers of detected features, green circles are selected features. b) Circles corresponding to selected features, circles' radii are proportional to features' scales.*

This great number of accumulated features consequently leads to a high time demanding classification procedure and thus the training data should be intelligently processed.

Among extracted features some are outliers that either were by accident wrongly hand-segmented, or are non-informative and represent statistically very improbable patches of texture. To omit these outliers, features from each terrain class are processed as follows:

1) Calculate distance matrix D, which contains distances between all pairs of descriptors:

$$D = \begin{bmatrix} d_{1,1} & \cdots & d_{N,1} \\ & & \\ d_{1,N} & \cdots & d_{N,N} \end{bmatrix} \quad (2)$$

2) Sort up each row of matrix D and add up first *k* elements with lowest values:

$$D' = \text{sort}(D) \quad (3)$$

$$\mu_j = \sum_{i=1}^{k} e^{-\frac{D'^2_{j,i}}{\sigma}} \quad (4)$$

Eliminate 5-20% of total number of descriptors with smallest $\mu$.

After eliminating 10% of the features in each class, the total number of remained features is 10746.

Before we proceed with the description of machine learning algorithms, it is useful to visualize feature space. This visual information allows us to understand how features of different classes are scattered, which is useful for parameters selection in classification routines. A number of approaches have been proposed to visualize high dimensional data. One of them is based on dimension reduction techniques. High-dimensional feature vectors are transformed into vectors with three components, so they can be plotted in the three dimensional space. We applied two approaches to reduce feature space dimensions that do not take into account the information on features' labels. The first one is a linear technique based on principal component analysis (PCA). PCA performances linear mapping in a way that variance of the data in the dimensionally reduced space is maximized. The reduced feature space is shown in figure 6a. Another approach for feature dimension reduction consists in applying a multi-layer perceptron (MLP) with a bottle neck principle. The number of input units and neurons in the output layer are set equal to the number of feature dimensions. The number of neurons in the hidden layer is set to a desired lower number, which represents the dimension of the reduced feature space. In our case the MLP structure is 64-3-64. The MLP learns to compress the data from 64 to 3 and then back to 64 dimensions. After the training process is finished, the outputs of neurons in the hidden layer represent new low-dimensional feature vectors (fig. 5d). For further analysis we separate a feature set, in the original 64-dimensional, space into two subsets. The first subset combines features whose $N$ nearest neighbors are features of the same class (fig. 5b, 5e), and the second subset contains the remaining features (fig. 5c, 5f).

Points from the first subspace are located deep inside clusters, far from cluster boundaries and therefore are less informative. The algorithm that separate one subset from the other can be summarized in the list of steps as follows

1) Calculate distance matrix D, which contains distances between all pair of descriptors;
2) Sort up each row of matrix D and choose first N nearest neighbors;
3) If all of them belong to the same class, then the feature is placed into the subset with dense features otherwise it is placed in the subset with non-dense features.

When $N = 5$, 39% of features fell in the first dense subset, and the remaining 61% is within the second subset. This ratio supports the assumption that there is an underline structure presented in the features space, meaning that features corresponding to similar textures are also located in close proximity. Furthermore, it can be seen from figure 6 that both linear and non-linear dimension reduction techniques generate very similar feature distributions. Therefore, the data can be separated by a function with less parameters than the number of features in the training set. It is interesting to notice that blue features representing trees are distributed in two groups. This fact is due to two visually different groups of trees. The first group contains trees with crowns, so that textures of leaves are distinguishable. The second group is trees without a crown, so that tree branches are visible. Difference in appearance of leaves and branches leads to two separable clusters.

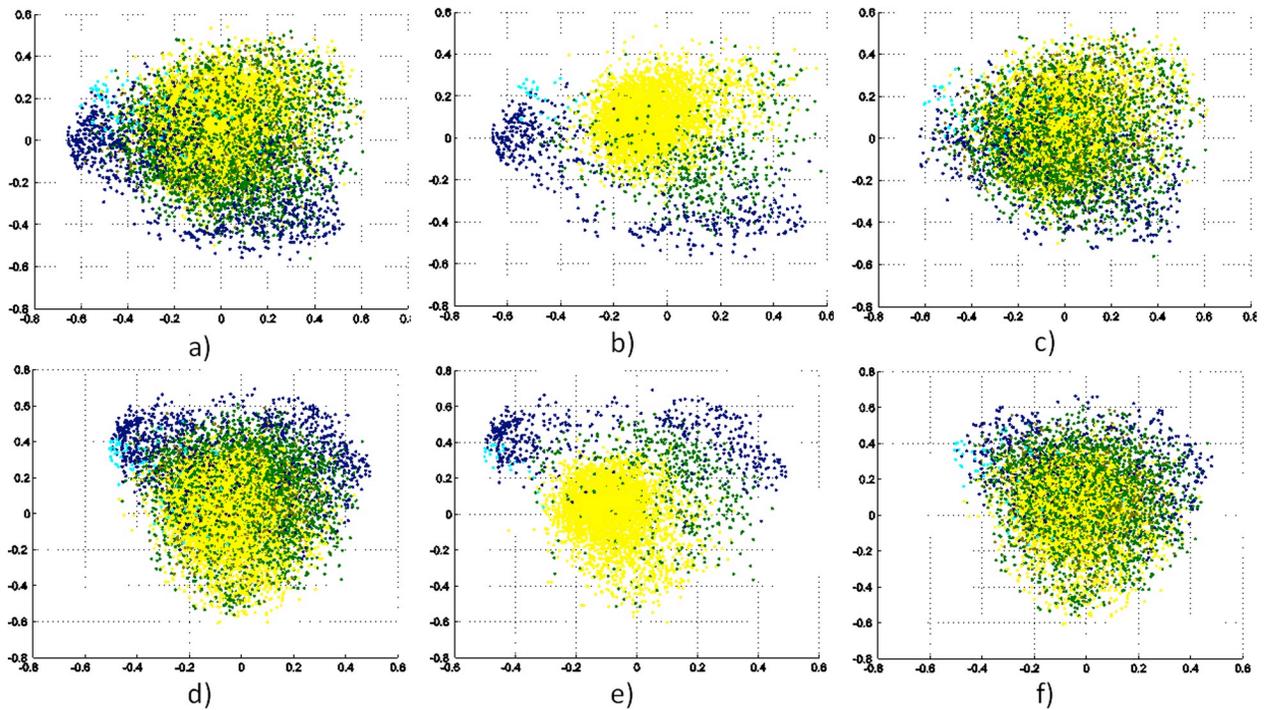

*Figure 5. Reduced feature space using PCA a) full set, b) dense features subset, c) non-dense features subset, and using MLP d) full set, e) dense features subset, f) non-dense features subset.*

## MACHINE LEARNING ALGORITHMS

Visual analysis of figure 5 suggests that some features are more discriminative than the others. It is crucial to select those features which contribute to class partitioning more and omit those which are less informative. The benefits are a lower system complexity and less storage requirement. Moreover, it improves the classifier performance and reduces computation time.
In the next sections we consider three supervised classifiers. The purpose of a classifier is to predict or estimate the likeliness of class label of an input feature vector after having seen a number of training examples. A broad range of classifiers have been proposed. We consider the following widely used classifiers: *k*-Nearest Neighbor Algorithm, Multilayer Perceptron and Support Vector Machine.

### *k*-Nearest neighbor algorithm and Kernel method

In this section we discuss a modified lazy-learning algorithm which is based on the combination of two non-parametric estimators. The first is called the *k*-nearest neighbor algorithm and the second is the Parzen window method. Both algorithms are similar although with some differences. *k*-Nearest Neighbor is well developed machine learning algorithms with a long history (Cover & Hart, 1967). The algorithm is based on the assumption that feature vectors of the same class located in close proximity to each other. Then an unclassified vector can be classified by observing the class labels of nearest neighbors. It was shown that *k*-nearest neighbor is guaranteed to approach the Bayes error rate, for some value of *k*, where *k* increases as a function of the number of data points.

Let among $k$ nearest vectors $k_m$ are from class $c^{(m)}$ and let the total number of vectors in class $c^{(m)}$ be $n_m$. Then the class conditional probability $\hat{p}(x|c^{(m)})$ is

$$\hat{p}(x|c^{(m)}) = \frac{k_m}{n_m V}, \qquad (5)$$

$V$ is a volume centered at a point $x$, and the prior probability $p(c_m)$ defined as

$$\hat{p}(c^{(m)}) = \frac{n_m}{n}, \qquad (6)$$

$$n = \sum_{m=1}^{M} n_m \qquad (7)$$

The class of an unclassified point $x$ is determined using the following rule:

$$c = \arg\max_{i=1..M} \hat{p}(c^{(i)}|x) \qquad (8)$$

Or applying Bayes' theorem,

$$c = \arg\max_{i=1..M} \left( \frac{k_i}{n_i V} \cdot \frac{n_i}{n} \right) \qquad (9)$$

It can be easily seen from (9), that $x$ is assigned to the class with the largest number samples among $k$ nearest neighbors.

In the Parzen method (Parzen, 1962) instead of fixing the number of nearest neighbors the algorithm fixes the volume around the vector to be classified. Then the number of neighbors residing within the volume is counted and the class probability $\hat{p}(x|c^{(m)})$ is estimated as:

$$\hat{p}(x|c^{(m)}) = \frac{1}{n_m h^p} \cdot \sum_{i=1}^{n_m} K\left( \frac{x - x_i^{(m)}}{h} \right) \qquad (10)$$

Where $K$ is a kernel function and $h$ is the smoothing parameter. To make $\hat{p}(x)$ satisfy properties of probability density function $p(x) \geq 0$ and $\int p(x) dx = 1$, kernel $K$ should conform to the following conditions: $K(z) \geq 0$ and $\int_{R^p} K(z) = 1$. The most popular kernel is the Gaussian kernel:

$$K(z) = \frac{1}{\left(\sqrt{2 \cdot \pi}\right)^p} \cdot \exp(-\frac{x^T x}{2}), \qquad (11)$$

where $p$ is a number of dimensions.

With estimated class PDFs the following classification rule can be used to classify a new feature:

$$c = \arg\max_{i=1..M} \hat{p}\left(x|c^{(m)}\right). \tag{12}$$

The algorithm we use in our experiments is based on Parzen window method the difference is that it estimates local probability density function by taking into account only *k*-nearest neighbors. This approach allows us to save time on computing many Gaussian kernels in dense regions or to take into account *k*-nearest neighbors in sparse regions.

*k*-Nearest Neighbor as well as Parzen window methods are both classified as instance-based learning algorithms. Instance-based learning algorithms delay their generalization process until the classification stage. This aspect leads to large memory requirements due to the necessity of storing all training samples. A consequence of large training sets is high computational demands related to calculating distances to each training sample. Another disadvantage particularly related to *k*-NN and Parzen methods is a requirement of choosing the similarity function as well as *k* or *h* parameters.

**Multi-layer perceptron**

Another approach to decrease calculation time and increase generalization of the segmentation system consists in applying a classifier based on a multilayer feed-forward neural network or more specifically on the multilayer perceptron(MLP). A classifier based on MLP has a few advantages against the lazy learning algorithms discussed early. Instead of computing distances between an input vector and all features from the training set, the MLP learns to transform training vectors into matrices of interlayer weight coefficients. The total number of coefficients is usually substantially less than the number of training samples multiplied by the number of components in a feature vector. As a consequence fast computation can be achieved with less memory requirement. The training process consists in turning coefficients of interlayer $W^{Hi}$ and output $W^O$ matrices. In our experiments we used a neural network with two hidden layers. It has been proved that an MLP with one hidden layer with sigmoid activation functions is capable of approximating any function with multiple arguments to arbitrary accuracy (Cybenko, 1989). However, for classification purposes, an MLP has to have two hidden layers to be able to separate concave domains. The decision function of an MLP with two hidden layers with sigmoid activation functions and with linear action function in output neurons (fig. 6) can be written as follows:

$$y_l = \sum_{i=0}^{N3} w_{l,i}^{(3)} f\left(\sum_{j=0}^{N2} w_{i,j}^{(2)} f\left(\sum_{k=0}^{N1} w_{j,k}^{(1)} x_k\right)\right) \tag{13}$$

*x* is an input SURF vector and *f* is a sigmoid function. We used the following hyperbolic sigmoid function:

$$f(x) = \frac{2}{\left(1+e^{-2\cdot x}\right)} - 1 \tag{14}$$

The training set contains pairs of input $x$ and known target vectors $t$. The number of network inputs corresponds to the number of dimensions of a SURF vector and the number of outputs is equal to the number of classes. The target vector is filled with -1 for all elements except the one $\hat{p}$ representing the class of input vector which equals 1. A feature vector $v$ is classified by simply choosing the class $c$ with maximum output:

$$c = arg\left(max_{1 \leq l \leq 5}(y_l)\right) \tag{15}$$

Our image segmentation algorithm takes into account not only the class label but also the likeliness of belonging to that as well as to other classes. Therefore, in the image segmentation process the whole vector $y$ is used. Although negative output vector components are set to zero.

The goal of the learning procedure is to find weights $W$ that minimize the following criterion obtained by substituting (3) into (1):

$$\varepsilon(w) = \sum_{m=1}^{M}\left(\sum_{i=0}^{N3} w_{l,i}^{(3)} f\left(\sum_{j=0}^{N2} w_{i,j}^{(2)} f\left(\sum_{k=0}^{N1} w_{j,k}^{(1)} x_k\right)\right) - t_m\right)^2 \rightarrow min \tag{16}$$

The most popular method for learning in multilayer networks is called Back-propagation. The idea behind the learning algorithm is the repeated application of the chain rule which allows finding how much each of the weights contributes to the network error (17):

$$\frac{\partial \varepsilon}{\partial w_{i,j}} = \frac{\partial \varepsilon}{\partial f_i} \frac{\partial f_i}{\partial u_i} \frac{\partial u_i}{\partial w_{i,j}}, \tag{17}$$

Then according to calculated errors modify each weight (18) to minimize the error:

$$w_{i,j}(t+1) = w_{i,j}(t) - \mu \frac{\partial \varepsilon}{\partial w_{i,j}}(t). \tag{18}$$

The ordinary gradient decent by back propagation is slow and often ends far from the optimal solution. To improve the quality of minimum search a number of modifications have been proposed. One is called RPROP, or 'resilient propagation' (Riedmiller & Braun, 1993). The idea behind the algorithm is to introduce for each weight its individual update-value $\Delta_{i,j}$ which determines the size of the weight-update. Every time the partial derivative of the corresponding weight $w_{i,j}$ changes its sign, which indicates that the last update was too big and the algorithm has jumped over a local minimum, the update-value $\Delta_{i,j}$ is decreased by the factor $\mu^-$. If the derivative retains its sign, $\Delta_{i,j}$ is slightly increased in order to accelerate convergence in shallow regions. After all update-values are adapted, neural weights are adjusted as follows:

$$\Delta w_{i,j}(t) = \begin{cases} -\Delta_{i,j}(t) & \text{if } \dfrac{\partial \varepsilon}{\partial w_{i,j}}(t) > 0 \\ +\Delta_{i,j}(t) & \text{if } \dfrac{\partial \varepsilon}{\partial w_{i,j}}(t) < 0, \\ 0 & \text{else} \end{cases} \quad (19)$$

$$w_{i,j}(t+1) = w_{i,j}(t) - \Delta w_{i,j}(t). \quad (20)$$

Probably the most successful and widely used learning algorithm is the Levenberg–Marquardt algorithm (Martin Hagan & Menhaj, 1994). The quasi-Newton methods are considered to be more efficient than gradient decent methods, but their storage and computational requirements go up as the square of the size of the network. Levenberg–Marquardt algorithm (LMA) is taking advantages of both Gauss–Newton algorithm and the method of gradient descent.
If error function is simply written as

$$\varepsilon(w) = \sum_{m=1}^{M} (e_m(w))^2 \quad (21)$$

and

$$e_m(w) = y_i(w) - t_i \quad (22)$$

then using the following notation:

$$e(w) = \begin{bmatrix} e_1(w) \\ e_2(w) \\ \ldots \\ e_M(w) \end{bmatrix}, \quad J(w) = \begin{bmatrix} \dfrac{\partial e_1}{\partial w_1} & \dfrac{\partial e_1}{\partial w_2} & \cdots & \dfrac{\partial e_1}{\partial w_n} \\ \dfrac{\partial e_2}{\partial w_1} & \dfrac{\partial e_2}{\partial w_2} & \cdots & \dfrac{\partial e_2}{\partial w_n} \\ \ldots & \ldots & \ldots & \ldots \\ \dfrac{\partial e_M}{\partial w_1} & \dfrac{\partial e_M}{\partial w_2} & \cdots & \dfrac{\partial e_M}{\partial w_n} \end{bmatrix} \quad (23)$$

Gradient vector and Hessian approximation corresponding to (21) defined as

$$g(w) = [J(w)]^T e(w), \quad (24)$$

$$G(w) = [J(w)]^T J(w) + R(w), \quad (25)$$

where $R(w)$ contains Hessian components of higher order derivatives.

The main idea of the LMA consists in approximating $R(w)$ with regularized parameter $v_k I$, so that Hessian is approximated as follows:

$$G(w_k) = [J(w_k)]^T J(w_k) + v_k I. \tag{26}$$

Then at the beginning of learning procedure, when $w_k$ is far from the optimal solution, $v_k$ is substantially higher than eigenvalues of $[J(w_k)]^T J(w_k)$. In this situation the Hessian matrix is replaced with:

$$G(w_k) = v_k I, \tag{27}$$

and minimization direction is chosen using the method of gradient descent:

$$p_k = -\frac{g(w_k)}{v_k}. \tag{28}$$

However, while the error is reducing the parameter $v_k$ is reducing too and therefore the first component in (26) start contributing more and more. Then $v_k$ is close to zero, the equation (26) is turning into the Gauss–Newton algorithm.

One of the advantages of the LMA is that it converges in less number of iterations than when resilient propagation is used.

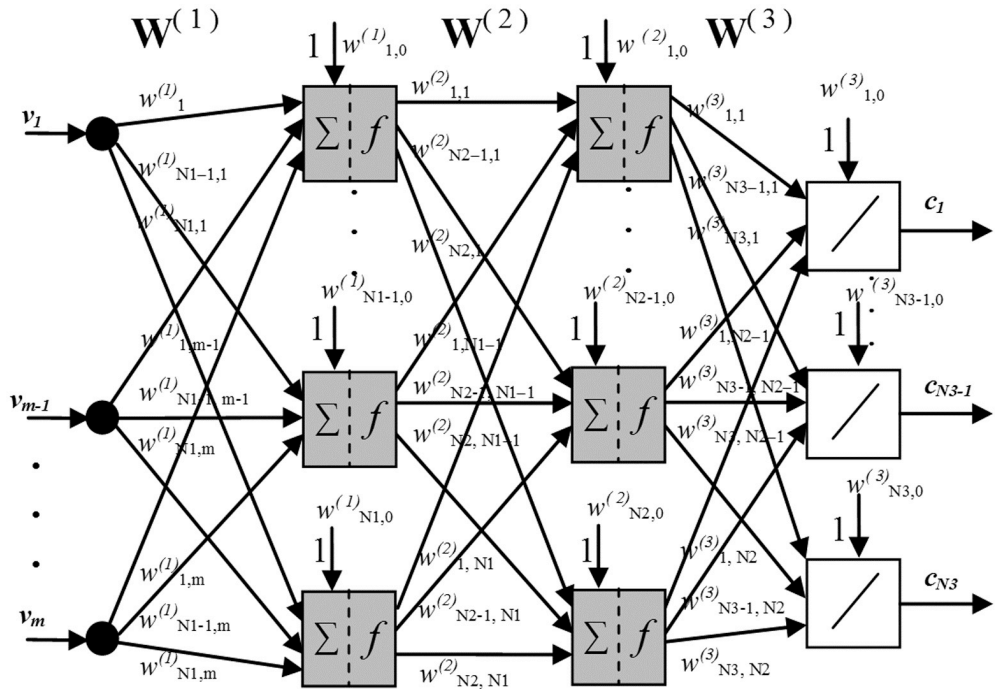

*Figure 6. Three layer perceptron*

**Support vector machine**

Support Vector Machines are relatively new machine learning algorithm (Vapnik, 1995). They transform a classification problem into quadratic programming problem, which always allows us to find an optimal solution. An advantage of SVMs consists in a good ability to separate very complex domains due to their ability of nonlinearly transforming data into a higher dimensional space, where hyper planes can separate already lineralized data (Fig 7). Another advantage is that SVMs selects only those vectors which are located close to the class boundary (support vectors) and therefore reduce the number of features.

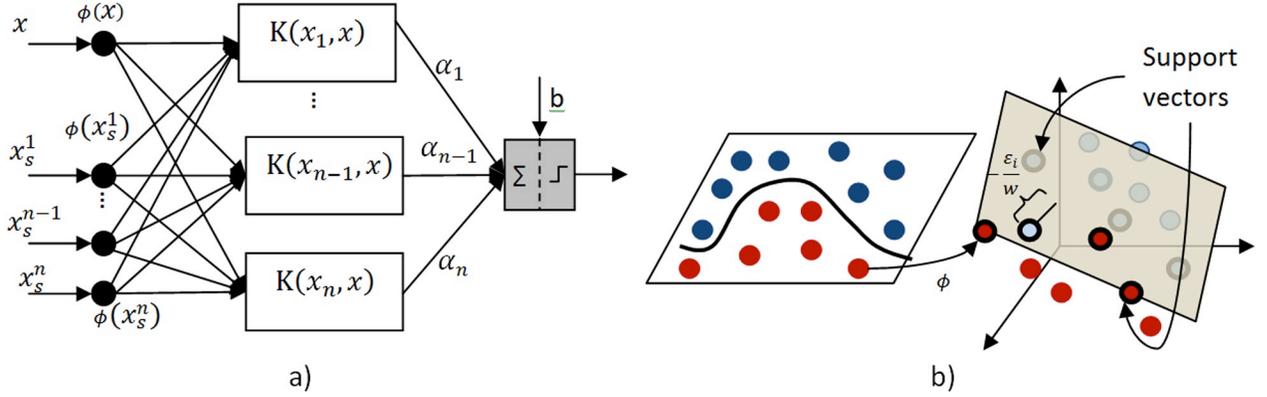

Figure 7. a) Block scheme of the SVM decision function, b) illustration of nonlinear transform from 2D to 3D space, bold circles highlight support vectors.

The decision function for the Kernel-SVM is:
$$f(x)=\sum_i \alpha_i K(x_i,x)+b \qquad (29)$$

$K(x_i,x)$ is the kernel function of the following form:
$$K(x_i,x)=\Phi(x_i)\cdot\Phi(x) \qquad (30)$$

To tune the SVM parameters the following optimization problem should be solved:
$$min_\alpha D(\alpha)=\frac{1}{2}\sum_{i,j}\alpha_i\alpha_j\Phi(x_i)\cdot\Phi(x_j)-\sum_i y_i\alpha_i \quad s.t. \begin{cases}\sum_i \alpha_i = 0 \\ 0\leq y_i\alpha_i \leq C\end{cases} \qquad (31)$$

As a kernel function we used RBF kernel:
$$K(x,x')=\exp(-\gamma x-x'^2) \qquad (32)$$

For multi-class classification we followed one-against-one approach, when $k(k-1)/2$ binary classifiers are constructed and each one trains data from two different classes. To decide to which class an input vector belongs to, each binary classifier is considered. A vector is classified as the class with the maximum number of assignments among $\frac{k(k-1)}{2}$ classifiers.

There are two parameters $C$ the upper bound and $\gamma$ the Kernel's width which greatly influence the classification accuracy. To select the optimal parameters we ran a set of experiments for different values of $C$ and $\gamma$. We calculated classification accuracy for each pair of the parameters. The classification accuracy is calculated as follows:

$$Accuracy = \frac{\#correctly\ predicted\ data}{\#total\ data} 100\% \qquad (33)$$

Parameters that yield the best accuracy are selected for the model construction.

| $\gamma \backslash C$ | $2^{-1}$ | $2^0$ | $2^1$ | $2^2$ | $2^3$ | $2^4$ | $2^5$ | $2^6$ | $2^7$ |
|---|---|---|---|---|---|---|---|---|---|
| $2^{-4}$ | 62.73 | 64.39 | 65.44 | 66.12 | 66.49 | * | * | * | * |
| $2^{-3}$ | 64.50 | 65.55 | 66.25 | 66.79 | 68.65 | * | * | * | * |
| $2^{-2}$ | 65.89 | 66.55 | 67.33 | 67.86 | 68.65 | * | * | * | * |
| $2^{-1}$ | 67.02 | 67.90 | 68.55 | 69.38 | 69.80 | 70.89 | 71.54 | 72.03 | * |
| $2^0$ | 68.30 | 69.12 | 69.80 | 71.02 | 71.74 | 72.28 | 72.63 | 72.52 | * |
| $2^1$ | 69.46 | 70.78 | 71.42 | 72.49 | 73.07 | 73.19 | 72.54 | 71.84 | * |
| $2^2$ | * | 72.18 | 73.14 | 73.6 | 73.43 | 72.35 | 71.51 | 70.82 | * |
| $2^3$ | * | 73.45 | 74.08 | **74.12** | 73.02 | 72.53 | 72.48 | 72.49 | 72.49 |
| $2^4$ | * | 72.87 | 73.97 | 73.56 | 73.59 | 73.62 | 73.62 | 73.62 | 73.62 |
| $2^5$ | * | * | 65.51 | 65.1 | 65.1 | 65.53 | 65.53 | 65.43 | 65.53 |

*Table 1. Parameters selection. Star sign indicates that particularly pair of parameters was not checked.*

In our image segmentation algorithm it is necessary to know the probabilities of an input vector belonging to each class. The probabilities are estimates as described in (Chang & Lin, 2010).

**EXPERIMENTS AND DISCUSSION**

The experiments were conducted for the training and testing datasets. First we started with the lazy learning algorithm and applied it to a set of different salient features. We tested the classifier on two datasets, one constructed of 64-dimesnional SURF features and the second consists of upright SURF features i.e. without rotational invariance. Comparing the error rates for these two datasets, and found that the segmentation error rate is much lower for U-SURF features. This result coincides with the results obtained for Gabor features in (Castano, et al., 2001). Gabor features without rotational invariance showed better performance than those with rotational invariance. The advantage of U-SURF features compared to SURF features is that the former is faster to compute and they also increase class distinctivity, while maintaining robustness to rotation of about ±15°. Therefore, for further analysis only variations of U-SURF features are considered. To decrease memory demands and decrease computation time we looked into feature vector reduction. One approach to reduce dimensionality is to use separately sums of either $d_y$ and $d_x$ (USURF32) or of their absolute values $|d_x|$ and $|d_y|$ (USURF32abs). In this case, the feature's dimension reduces by half and equals to 32. Another way consists in reducing the quadratic grid laid over the interest point from 4 x 4 to 3 x 3. Then, the number of dimensions reduces to 36. Among these three dimensionally reduced features, the best performance is achieved with 36 dimensional vectors. Surprisingly, the 36-dimension version of U-SURF performs much better on the test data than in the case of 64-dimension U-SURF descriptors. For that reason in further experiments we use the upright version of SURF features with 3 x 3 quadratic grid. We also experimented with different numbers of neighbors, specifically with $k = \{1,2,3,4,5\}$. The parameter $k$ did not greatly influence the error rate for the training data set as well as the testing data, although when the parameter $k=3$ the algorithm performed slightly better. Finally, we divided the U-SURF36 set into the subsets of dense and non-dense features as discussed earlier and then estimated error rates for both of them for $k = 1$ and $k = 3$. The total number of features was 2823 and 7922 for dense and non-dense feature subsets correspondingly. The error rates of dense features data significantly increased, however for non-dense features the results are comparable to the above discussed feature sets for both training and testing image sets.

We conducted three experiments with the classifier based on a neural network. In the first and second experiments we applied the Levenberg–Marquardt learning algorithm for two network structures. In both cases the training set consisted of U-SURF 36 dimensional vectors. Half of the set was used for training, a quarter for testing, and another quarter for validation. The structure of the first network was 36-40-40-5 and of the second was 36-60-60-5. In the third experiment the latter network structure was used; however, the learning algorithm was switched to resilient propagation. The experimental results show that a neural network with the structure of 36-40-40-5 is less effective than the network with 36-60-60-5. Among learning algorithms, the network trained with LMA shows the lowest error rate. It is worth noting that in terms of memory and computation requirement neural network substantially outperforms lazy learning algorithms. In the case when the network with a structure of 36-60-60-5 is used, the number of coefficients stored in the network is (36 x 60 + 60) + (60 x 60 + 60) + (60 x 5 + 5) = 6185, but in the case when lazy learning is used the number of coefficients is 10746 x 36 = 386856 for all training vectors and 7922 x 36 = 285192 for only none-dense vectors. So the number of coefficients necessary for classification is based on neural network constitutes for (6185/386856) * 100% = 1.6% of the whole training set and (6185/285192) * 100% = 2.1% for non-dense features. Nevertheless, a neural network classifier achieves comparable results.

The last classifier we experimented with was an SVM. The choice of parameters for the SVM greatly impacts the classification accuracy. Performing parameters selection procedure as discussed early in table 1, we found that the best accuracy is achieved for $\gamma = 8$ and $C = 4$. For these parameters 8101 support vectors were selected. Which constitutes for 8101/10746 * 100% = 75% of the entire training set.

The error rates for all the above discussed classifiers and salient features are presented in tables 2 and 3. In the left column an abbreviation for classifier and type of salient features are given. SURF and USURF stands for rotationally variant and invariant features. The number followed after SURF indicate vectors dimension. NN1 to NN5 stands for abbreviation of nearest number and the number correspond to the number of neighbors which were considered during the classification. DF and NDF stands for dense and non-dense feature sets. MLP40, MLP60 stands for multilayer perceptron with 36-40-40-5 and 36-60-60-5 structures trained with LMA and MLP60RP is MLP trained with RPPROP. In the central parts error rates for ten images are given. The right column contains an average error rate among 10 images and its standard deviation.

| | | | | | | | | | | | |
|---|---|---|---|---|---|---|---|---|---|---|---|
| SURF64 | 6.03 | 4.78 | 3.90 | 4.61 | 4.10 | 7.74 | 6.10 | 6.33 | 6.50 | 5.20 | 5.53±1.21 |
| USURF64 | 6.77 | 4.90 | 4.03 | 4.86 | 4.34 | 8.27 | 6.88 | 6.61 | 6.28 | 5.38 | 5.83±1.34 |
| USURF32 | 6.40 | 4.95 | 3.82 | 4.81 | 4.69 | 7.39 | 6.90 | 6.511 | 6.44 | 5.29 | 5.72±1.15 |
| USURF32abs | 9.09 | 5.89 | 6.45 | 6.35 | 5.62 | 10.08 | 7.25 | 9.11 | 8.36 | 6.12 | 7.43±1.59 |
| USURF36NN1S1 | 5.39 | 4.82 | 3.73 | 4.55 | 4.12 | 6.90 | 5.79 | 6.03 | 6.03 | 21.08 | 6.84±5.09 |
| USURF36NN1 | 5.73 | 4.84 | 4.01 | 4.57 | 4.08 | 7.27 | 6.90 | 6.43 | 6.67 | 5.27 | 5.58±1.19 |
| USURF36NN2 | 5.93 | 4.89 | 4.08 | 4.57 | 4.20 | 7.33 | 7.85 | 6.63 | 6.86 | 5.25 | 5.76±1.35 |
| USURF36NN3 | 6.05 | 4.95 | 4.07 | 4.84 | 4.23 | 7.48 | 8.86 | 6.80 | 7.01 | 5.44 | 5.97±1.54 |
| USURF36NN4 | 6.10 | 5.07 | 4.13 | 4.96 | 4.24 | 7.66 | 10.03 | 7.00 | 7.16 | 5.64 | 6.20±1.81 |
| USURF36NN5 | 6.20 | 5.14 | 4.19 | 5.04 | 4.30 | 7.93 | 11.27 | 7.27 | 7.29 | 5.87 | 6.45±2.12 |
| USURF36DF1 | 24.34 | 21.70 | 15.54 | 14.54 | 13.59 | 25.01 | 32.93 | 20.37 | 18.46 | 22.00 | 20.85±5.81 |
| USURF36DF3 | 26.49 | 22.25 | 16.42 | 14.53 | 14.83 | 26.48 | 33.53 | 21.96 | 18.85 | 23.07 | 21.84±5.98 |
| USURF36NDF1 | 6.27 | 4.79 | 4.24 | 4.85 | 4.28 | 7.22 | 7.16 | 7.24 | 6.82 | 5.80 | 5.87±1.23 |
| USURF36NDF3 | 6.08 | 4.86 | 4.19 | 4.98 | 4.32 | 7.39 | 9.08 | 7.28 | 7.05 | 6.24 | 6.15±1.57 |
| USURF36MLP40 | 23.83 | 24.53 | 18.94 | 13.45 | 12.20 | 23.01 | 32.08 | 21.02 | 22.29 | 21.49 | 21.28±5.64 |
| USURF36MLP60 | 21.40 | 20.65 | 14.11 | 12.32 | 10.60 | 23.21 | 29.53 | 18.75 | 16.55 | 19.69 | 18.68±5.59 |
| USURF36MLP6RP | 25.83 | 22.81 | 17.28 | 13.82 | 12.40 | 24.83 | 31.22 | 18.93 | 19.25 | 21.41 | 20.78±5.69 |
| USURF36SVM | 7.05 | 6.02 | 3.64 | 4.90 | 4.65 | 7.89 | 13.57 | 7.351 | 7.13 | 6.70 | 6.89±2.71 |

*Table 2. Experimental results conducted on training dataset.*

| | | | | | | | | | | |
|---|---|---|---|---|---|---|---|---|---|---|
| SURF64 | 15.61 | 23.91 | 18.20 | 20.21 | 23.31 | 25.13 | 20.23 | 17.80 | 19.16 | 19.49 | 20.31±2.98 |
| USURF64 | 16.03 | 31.60 | 22.39 | 21.68 | 26.09 | 32.44 | 25.12 | 20.29 | 21.97 | 25.57 | 24.32±5.02 |
| USURF36NN1S1 | 11.28 | 19.93 | 15.75 | 17.27 | 19.04 | 19.34 | 17.12 | 16.61 | 18.48 | 22.17 | 17.70±2.92 |
| USURF32 | 16.96 | 24.61 | 19.18 | 22.46 | 21.89 | 24.60 | 22.66 | 19.94 | 21.38 | 23.38 | 21.71±2.43 |
| USURF32abs | 20.94 | 29.06 | 23.08 | 21.79 | 28.69 | 33.46 | 26.77 | 25.56 | 24.08 | 20.35 | 25.38±4.17 |
| USURF36NN1 | 13.31 | 19.53 | 16.32 | 17.45 | 20.18 | 20.66 | 18.42 | 17.66 | 18.66 | 17.55 | 17.97±2.11 |
| USURF36NN2 | 12.42 | 19.70 | 15.55 | 16.23 | 18.66 | 19.31 | 18.40 | 17.34 | 17.73 | 17.75 | 17.31±2.14 |
| USURF36NN3 | 12.48 | 19.76 | 15.24 | 15.91 | 18.09 | 19.22 | 18.63 | 17.69 | 17.57 | 18.15 | 17.27±2.17 |
| USURF36NN4 | 12.70 | 19.63 | 15.09 | 15.79 | 17.98 | 19.32 | 18.90 | 18.02 | 17.41 | 18.53 | 17.34±2.18 |
| USURF36NN5 | 13.13 | 19.75 | 15.03 | 15.74 | 17.85 | 19.58 | 19.29 | 18.35 | 17.33 | 18.91 | 17.50±2.20 |
| USURF36DF1 | 26.89 | 23.52 | 18.37 | 16.36 | 19.95 | 25.10 | 23.58 | 23.65 | 20.10 | 25.56 | 22.31±3.42 |
| USURF36DF3 | 27.77 | 24.72 | 19.33 | 16.52 | 20.46 | 26.44 | 24.13 | 24.05 | 20.28 | 25.87 | 22.96±3.61 |
| USURF36NDF1 | 15.30 | 22.20 | 19.06 | 19.59 | 22.71 | 21.67 | 19.18 | 19.07 | 20.97 | 19.35 | 19.91±2.13 |
| USURF36NDF3 | 13.51 | 21.07 | 16.79 | 16.91 | 19.78 | 19.37 | 18.63 | 18.08 | 18.24 | 19.15 | 18.15±2.07 |
| USURF36MLP40 | 30.59 | 23.56 | 14.50 | 16.50 | 22.71 | 24.97 | 22.68 | 24.35 | 21.67 | 21.53 | 22.31±4.43 |
| USURF36MLP60 | 25.46 | 23.30 | 15.61 | 14.38 | 20.17 | 23.67 | 21.62 | 22.39 | 20.40 | 22.32 | 20.13±3.50 |
| USURF36MLP6RP | 23.25 | 21.49 | 15.54 | 14.23 | 22.73 | 24.42 | 21.52 | 22.49 | 18.05 | 22.31 | 20.60±3.44 |
| USURF36SVM | 14.76 | 20.43 | 12.45 | 14.09 | 17.42 | 18.06 | 19.95 | 18.35 | 18.08 | 19.01 | 17.26±2.63 |

*Table 3. Experimental results conducted on testing dataset.*

Some examples of terrain segmentation are shown on figure 8. The same training image as well as the same test image for each classifier is shown for comparison purposes.

**CONCLUSIONS**

Among all the experiments we conducted, non-rotationally invariant 36-dimensional vectors significantly outperformed the 64-dimensional version of SURF. Among the experiments with various classifiers on the testing dataset, the lowest error rate was achieved with the SVM classifier and with $k$-NN classifier, when $k=3$. However, the number of support vectors smaller than in the complete training dataset, make the SVM more suitable for terrain segmentation. Multi-layer perceptron trained with LMA, showed higher error rates, although if we consider that only 50% of the training sets were used for training and the other half for validation and testing, the error segmentation rate of 20.13% is respectable compared to the 17.26% for SVM. Moreover, as it was mentioned above the number of coefficients needed for classification is almost 100 times less than in the case of the SVM classifier.

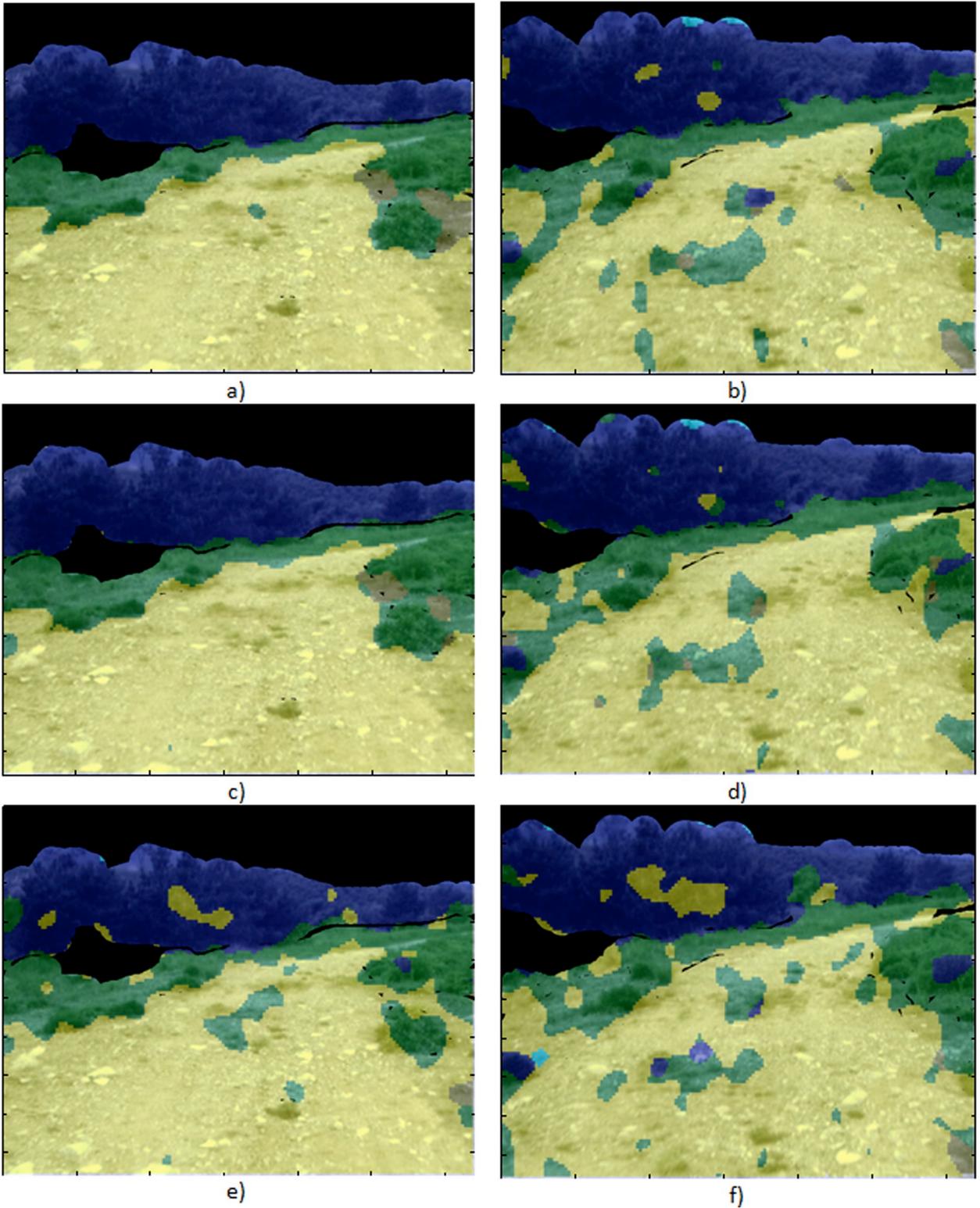

*Figure 8. Example of cross-country segmentation using SVM a), b), kNN with 3 neighbors c),d) and MLP with 60 neurons in each hidden layer e), f). The images a),c),e) are from the training image set and b),d),f) are from the testing set.*